\documentclass[letterpaper]{article} % DO NOT CHANGE THIS
\usepackage{aaai25}  % DO NOT CHANGE THIS
\usepackage{times}  % DO NOT CHANGE THIS
\usepackage{helvet}  % DO NOT CHANGE THIS
\usepackage{courier}  % DO NOT CHANGE THIS
\usepackage[hyphens]{url}  % DO NOT CHANGE THIS
\usepackage{graphicx} % DO NOT CHANGE THIS
\usepackage{natbib}  % DO NOT CHANGE THIS AND DO NOT ADD ANY OPTIONS TO IT
\usepackage{caption} % DO NOT CHANGE THIS AND DO NOT ADD ANY OPTIONS TO IT
\usepackage{algorithm}
\usepackage{algorithmic}
\usepackage{booktabs}
\usepackage{multirow}
\usepackage{amsmath}
\usepackage{amssymb}
\usepackage{appendix}
\usepackage{xcolor}

\usepackage{newfloat}
\usepackage{listings}
\DeclareCaptionStyle{ruled}{labelfont=normalfont,labelsep=colon,strut=off} % DO NOT CHANGE THIS
\floatstyle{ruled}
\newfloat{listing}{tb}{lst}{}
\floatname{listing}{Listing}
\title{3D Annotation-Free Learning by Distilling 2D Open-Vocabulary Segmentation Models for Autonomous Driving}
\author{
    Boyi Sun\textsuperscript{\rm 1, \rm 2},
    Yuhang Liu\textsuperscript{\rm 1, \rm 2},
    Xingxia Wang\textsuperscript{\rm 1}
    Bin Tian\textsuperscript{\rm 1, \rm 3}
    Long Chen\textsuperscript{\rm 1, \rm 3}
    Fei-Yue Wang\textsuperscript{\rm 1}\thanks{Corresponding Author.}
}
\affiliations{
    \textsuperscript{\rm 1}Institute of Automation, Chinese Academy of Sciences\\
    \textsuperscript{\rm 2}Zhongke JingYu Sensing Technology Co., Ltd\\
    \textsuperscript{\rm 3}Waytous\\
    \{sunboyi2024, liuyuhang2021, wangxingxia2022, bin.tian, long.chen, feiyue.wang\}@ia.ac.cn
}

\usepackage{bibentry}
\begin{document}

\maketitle

\begin{abstract}
Point cloud data labeling is considered a time-consuming and expensive task in autonomous driving, whereas annotation-free learning training can avoid it by learning point cloud representations from unannotated data. In this paper, we propose AFOV, a novel 3D \textbf{A}nnotation-\textbf{F}ree framework assisted by 2D \textbf{O}pen-\textbf{V}ocabulary segmentation models. It consists of two stages: In the first stage, we innovatively integrate high-quality textual and image features of 2D open-vocabulary models and propose the Tri-Modal contrastive Pre-training (TMP). In the second stage, spatial mapping between point clouds and images is utilized to generate pseudo-labels, enabling cross-modal knowledge distillation. Besides, we introduce the Approximate Flat Interaction (AFI) to address the noise during alignment and label confusion. To validate the superiority of AFOV, extensive experiments are conducted on multiple related datasets. We achieved a record-breaking 47.73\% mIoU on the annotation-free 3D segmentation task in nuScenes, surpassing the previous best model by 3.13\% mIoU. Meanwhile, the performance of fine-tuning with 1\% data on nuScenes and SemanticKITTI reached a remarkable 51.75\% mIoU and 48.14\% mIoU, outperforming all previous pre-trained models. 
% Our anonymous code is available at: https://anonymous.4open.science/r/AFOV2024.
Our code is available at: \url{https://github.com/sbysbysbys/AFOV}
% Our code is provided in the supplementary material and will be open-sourced.
\end{abstract}

\section{Introduction}
\label{sec:intro}

% ！！点云数据和图像数据，后面的图像数据要加上。
As neural-network-based 3D scene perception methods, \textit{e.g.}, object detection \citep{pointRCNN,pv-RCNN,voxelnet,pointpillar}, point cloud segmentation \citep{pointnet,pointnet++,Minkowski-net}, \textit{etc.}, become increasingly complex in their network architectures with a growing number of parameters, methods relying solely on enhancing model structures are reaching a point of saturation. Meanwhile, approaches to enhancing model performance through data-driven methods heavily rely on time-consuming and expensive manual annotations. Due to constraints such as insufficient class annotations, applying traditional point cloud perception methods to large-scale unlabeled data meets significant challenges.

\begin{figure}[t]
\centering
\includegraphics[width=\columnwidth]{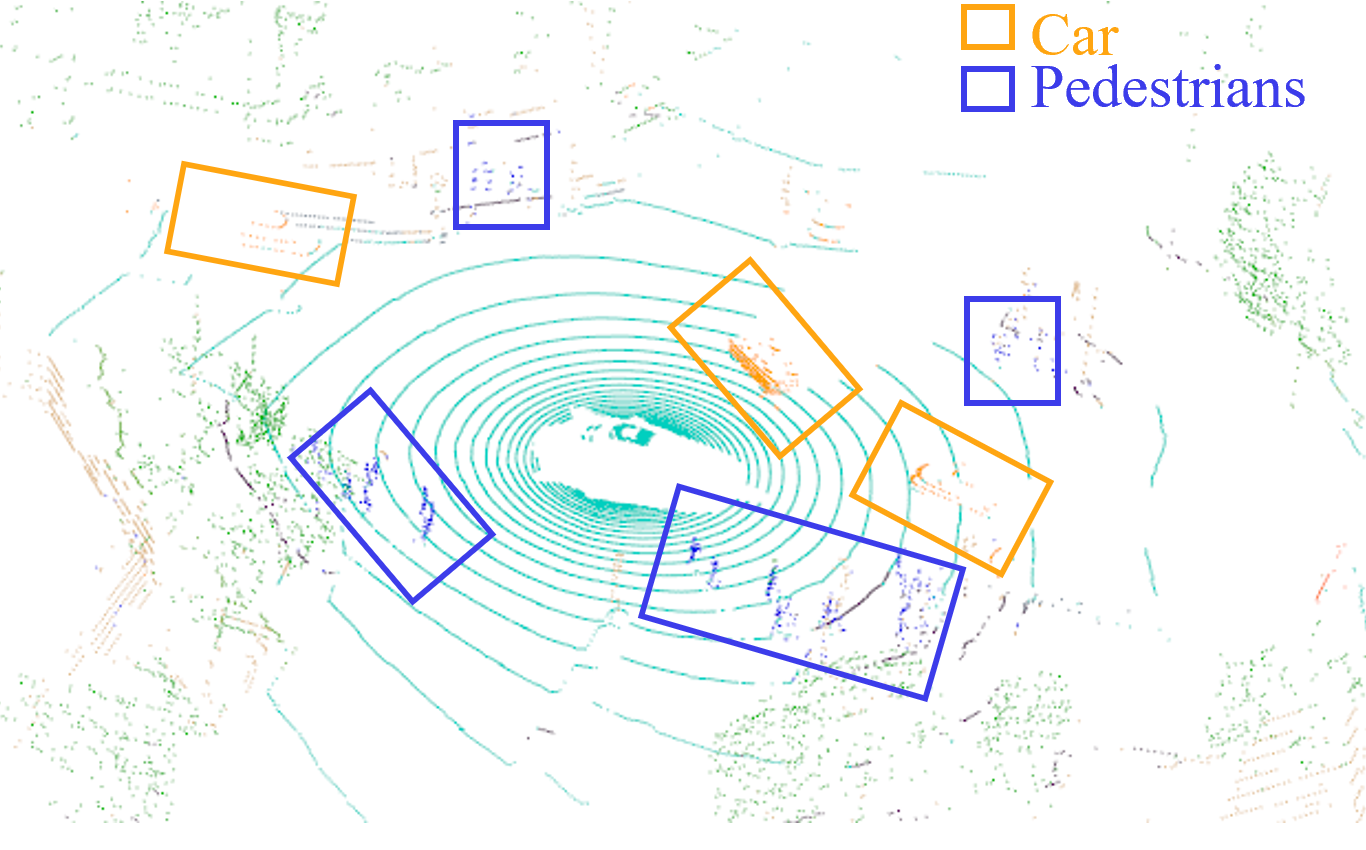}
\caption{Segmentation results of AFOV annotation-free training. More illustrations are presented in Appendix D.}
\label{fig1a}
\end{figure}

\begin{figure}[t]
\centering
\includegraphics[width=\columnwidth]{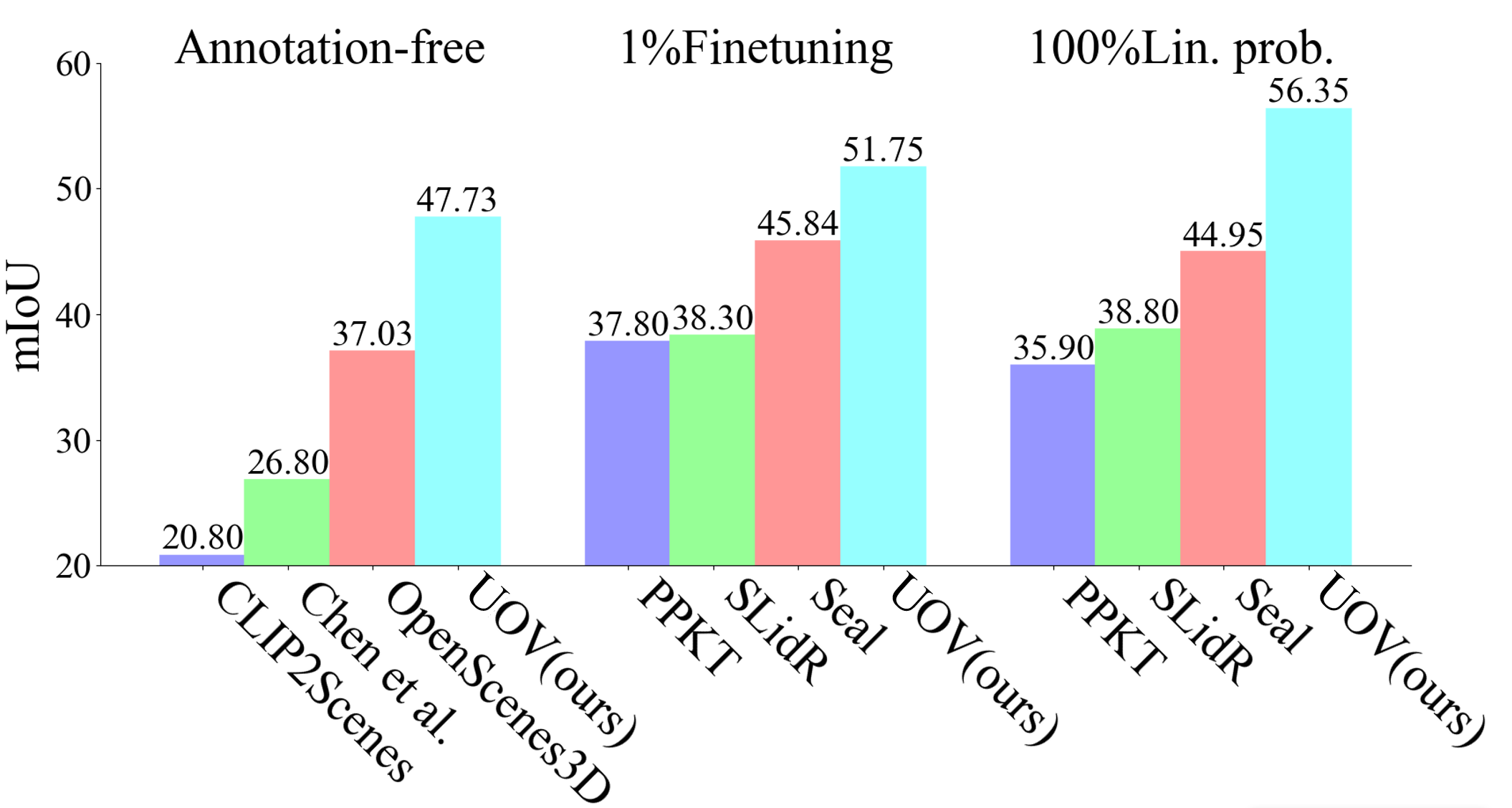}
\caption{Performance of AFOV on nuScenes.}
\label{fig1b}
\end{figure}

% \begin{figure}
%   \centering
%     \subfloat[Segmentation results of AFOV annotation-free training. More illustrations are presented in Fig. \ref{Appendix_B:performance_demo}]{
%     		\includegraphics[scale=0.35]{pic1_1.png}
%             \label{fig:a}
%             }
%     \subfloat[Results of various tasks on nuScenes]{
%     		\includegraphics[scale=0.29]{pic1_2.png}
%             \label{fig:b}}
%   \caption{Performance of AFOV on nuScenes \citep{nuscenes}. Our annotation-free segmentation approach demonstrates comparable results to some supervised methods (Fig. \ref{fig:a}), while achieving optimal performance across various tasks on nuScenes (Fig. \ref{fig:b}).}
%   \label{fig:performance}
% \end{figure}

Annotation-free learning is a powerful machine learning paradigm that enables representation learning from data without the need for manual supervision. Multi-modal annotation-free algorithms for 3D scene perception tasks integrate the internal structure of point clouds with image or text knowledge to generate objectives. They bypass the costly manual annotations, bridging the gap between traditional 3D perceptual models and unannotated data.

Existing 3D annotation-free methods \citep{CLIP2Scene,clip2,Towards_label-free,OpenScene} aim to transfer knowledge from visual foundation models (VFMs), \textit{e.g.}, Contrastive Vision-Language Pre-training (CLIP) \citep{clip} or Segment Anything (SAM) \citep{sam}, to point cloud representations. However, 3D annotation-free models based on CLIP \citep{clip} suffer from intolerable noise, while SAM \citep{sam} fails to correspond texts and images. Therefore, we seek high-quality image segmentation models with textual correspondences to serve as teacher models for 3D annotation-free learning. Recently, CLIP-based 2D open-vocabulary segmentation models  \citep{MaskCLIP,fcclip,san,CAT-Seg} have demonstrated excellent performances. They employ contrastive learning to extract textual and image features from a shared embedding space and are capable of segmenting and identifying objects from a set of open classes in various environments. These models provide us with image segmentation and labels corresponding to the segmented regions, as well as easy-to-extract text and image representations. Meanwhile, they significantly outperform other Visual Language Models (VLMs) \citep{clip,SEEM,X-Decoder,semantic-SAM} in open-vocabulary segmentation tasks.

In this paper, we propose AFOV, a 3D \textbf{A}nnotation-\textbf{F}ree framework by distilling 2D \textbf{O}pen-\textbf{V}ocabulary segmentation models. It aims to address the difficulties of point cloud perception using unannotated data through 3D annotation-free learning. 
AFOV adopts a novel two-stage strategy: The first stage uses Tri-Modal contrastive Pre-training (TMP) to warm up the network parameters, where we innovatively incorporate the textual information to enhance the semantic perception of point cloud representations; The second stage is pseudo-label guided annotation-free training, completing the knowledge distillation from 2D to 3D. Furthermore, to address the perceptual limitations of AFOV, such as noise during alignment and label confusion, we introduce Approximate Flat Interaction (AFI). It provides a robust error correction mechanism for the above two stages through point cloud spatial interaction.

In the experiment, We selected four CLIP-based open-vocabulary segmentation models, \textit{ie.}, MaskCLIP \citep{MaskCLIP}, FC-CLIP \citep{fcclip}, SAN \citep{san}, CAT-Seg \citep{CAT-Seg}, as teacher models. To validate the performance of our method in annotation-free point cloud segmentation tasks, extensive experiments on multiple autonomous driving datasets were conducted. Firstly, AFOV achieved a remarkable improvement of 3.13\% mIoU, reaching a Top-1 accuracy of 47.74\% mIoU in benchmark testing of annotation-free 3D segmentation on nuScenes \citep{nuscenes}. Furthermore, treating AFOV as a pre-trained model, we conducted 1\% data fine-tuning and 100\% data linear-probing experiments on nuScenes, yielding mIoU scores of 51.75\% and 56.35\%. Compared to the current best pretraining method, AFOV achieved improvements of 4.16\% and 4.81\% mIoU, respectively. When fine-tuning with 1\% data on SemanticKITTI (pre-training on nuScenes), AFOV achieved a 48.14\% mIoU. AFOV demonstrated state-of-the-art performance across various experiments, validating its effectiveness.

Compared to scene understanding work based on point clouds (such as OpenScene), the introduction of text information allows AFOV to generate pseudo labels for knowledge distillation. As a result, AFOV directly predicts outcomes, whereas scene understanding models match features between point clouds and text. As with fully supervised closed-set point cloud segmentation models, the directly predicted output is far superior to the output obtained through feature matching. 

Unlike most previous annotation-free 3D segmentation models, all the knowledge for training AFOV comes from state-of-the-art 2D open-vocabulary segmentation models. These open-vocabulary segmentation models (such as FC-CLIP, CAT-Seg, and SAN) perform far better than other VLMs like maskCLIP, SAM, and SEEM. They can also extract masks, labels, and features, avoiding noise accumulation between different backbones. Therefore, AFOV presents significant advantages over previous related works.

% AFOV employs a novel and efficient two-stage annotation-free training framework, which comprehensively utilizes state-of-the-art 2D open-vocabulary segmentation models for knowledge distilling. It innovatively introduces TMP and AFI, addressing the issues in previous works. Moreover, we introduce the superpixel-superpoint into annotation-free 3D segmentation for the first time. Experimentally, our approach not only breaks through in annotation-free semantic segmentation (Fig. \ref{fig1a}), but also notably outperforms prior state-of-the-art methods in other downstream tasks (Fig. \ref{fig1b}).

In conclusion, the main contributions of this work are:

\begin{itemize}

\item We introduce a novel and efficient two-stage annotation-free training framework, AFOV, which comprehensively utilizes state-of-the-art 2D open-vocabulary segmentation models for knowledge distilling.

\item AFOV innovatively introduces TMP and AFI, addressing the issues in previous works. Moreover, we introduce the superpixel-superpoint into annotation-free 3D segmentation for the first time.

\item Experimentally, our approach not only breaks through in annotation-free semantic segmentation (Fig. \ref{fig1a}), but also notably outperforms prior state-of-the-art methods in other downstream tasks (Fig. \ref{fig1b}).

\end{itemize}

% Overall, the key contributions of our work are summarized as follows:

% 1. Our approach comprehensively utilize state-of-the-art 2D open-vocabulary segmentation models for training 3D unsupervised networks. It is the first to use image-point cloud and text-point cloud contrastive learning based on superpoint-superpixel.

% 2. We propose a two-stage unsupervised framework, AFOV, which leverages 2D open-vocabulary segmentation models to assist in unsupervised training without annotation. Additionally, a detachable spatial interaction method, AFI, is introduced to enhance the spatial understanding of the model.

% 3. Our approach not only breaks through in annotation-free semantic segmentation (Fig. \ref{fig1a}), but also notably outperforms prior state-of-the-art methods in other downstream tasks (Fig. \ref{fig1b}).

\section{Related Work}

\subsection{CLIP-based 2D Open-Vocabulary Segmentation}

2D open-vocabulary segmentation models aim to segment all categories in the real world. Traditional open-vocabulary image segmentation models  \citep{originalov1,originalov2,originalov3} attempt to learn image embeddings aligned with text embeddings. Inspired by Visual Language Models (VLMs), \textit{e.g.}, CLIP \citep{clip} and ALIGN \citep{ALIGN}, which have demonstrated remarkable performance in 2D tasks, recent studies have attempted to transfer CLIP's outstanding zero-shot segmentation capability to open-vocabulary tasks \citep{san,CAT-Seg,MaskCLIP,fcclip,LSeg,openseg,zegformer,ZSseg,ovseg}. In notable works, LSeg \citep{LSeg} learns pixel-level visual embeddings from CLIP, marking the first exploration of CLIP's role in language-driven segmentation tasks. More recently, MaskCLIP \citep{MaskCLIP} obtains pixel-level embeddings by modifying the CLIP image encoder; SAN \citep{san} augments CLIP with lightweight side networks to predict mask proposals and categories; CAT-Seg \citep{CAT-Seg} proposes a cost-aggregation-based method to optimize the image-text similarity map. Additionally; FC-CLIP \citep{fcclip} integrates all components into a single-stage framework using a shared frozen convolutional CLIP backbone. These works utilize CLIP as the central component of their network, granting them robust segmentation and recognition capabilities, as well as an image-text-aligned structure. Consequently, they can provide us with high-quality knowledge.

\subsection{3D Representation Learning Based on 2D-to-3D Knowledge distillation}

Unsupervised learning can be utilized for learning point cloud representations. Mainstream 3D unsupervised pre-training methods are mainly reconstruction-based \citep{reconstruction_based1,reconstruction_based2,reconstruction_based3,reconstruction_based4}, or contrast-based \citep{contrast_based1,Pointcontrast,DepthContrast,PPKT,SLidR,ST-SLidR,seal}. However, many of these methods are constrained by the quantity of point clouds, limiting their applicability to single-object or indoor scene learning. Prior attempts, such as PointContrast  \citep{Pointcontrast}, DepthContrast  \citep{DepthContrast}, SegContrast  \citep{segcontrast}, and PPKT \citep{PPKT}, have built contrastive objectives on large-scale point clouds. Additionally, SLidR  \citep{SLidR} adopts a novel approach by leveraging a superpixel-superpoint correspondence for 3D-to-2D spatial alignment, showing promising performance on autonomous driving datasets. Built upon SLidR, SEAL \citep{seal} employs VLMs to aid in superpixel generation.

Recently, inspired by the achievements of CLIP \citep{clip}, numerous works have focused on reproducing the excellent performance demonstrated by CLIP in 3D annotation-free tasks, not limited to pre-training. In 3D scene understanding, CLIP2Scene \citep{CLIP2Scene}, similar to OpenScene \citep{OpenScene} and OV3D \citep{OV3D}, embeds the knowledge of CLIP feature space into representations of 3D point cloud, enabling annotation-free point cloud segmentation; PLA \citep{pla} and RegionPLC \citep{regionplc} accomplish scene understanding through point-language alignment or contrastive learning framework; VLM2Scenes \citep{vlm2scene} exploits the potential of VLMs; and CLIP2 \citep{clip2} demonstrates perfect zero-shot instance segmentation performance through language-3D alignment at the semantic level and image-3D alignment at the instance level. Unlike the others, Chen \textit{et al.} \citep{Towards_label-free} utilizes CLIP to generate pseudo-labels and uses SAM \citep{sam} to assist in denoising. 

% In general, most of these methods rely on either feature alignment or pseudo-labeling. However, using only contrastive learning may incorrectly pull apart positive samples of the same class when employing point-level correspondence methods, \textit{e.g.}, pixel-point, pixel to FPS and $k$-NN. It is also susceptible to differences in data volume for each class and asymmetric point cloud densities. Meanwhile, the use of pseudo-labeling is greatly limited by the accuracy of the teacher's model. To address these issues, our work employs state-of-the-art 2D open-vocabulary segmentation models as teacher models. We not only distill knowledge onto point cloud using pseudo-labels, but also incorporate multi-modal contrastive learning into our framework.

\section{Method}

\subsection{Extracting Knowledge from CLIP-based 2D Open-Vocabulary Segmentation Models}
\label{3.1}

In perceptual approaches to unknown classes in 2D, unlike zero-shot learning, open-vocabulary learning uses language data as supervision. In terms of network structure, MaskCLIP changes the image encoder of CLIP to propose pixel-level representations instead of image-level. SAN proposes a side adapter network attached to a frozen CLIP encoder; CAT-Seg employs a cost-aggregation-based method to improve CLIP; FC-CLIP adds a decoder, mask generator, in-vocabulary classifier, and out-vocabulary classifier after freezing the CLIP backbone. In most cases, the mask generator operates independently of the class generator. Pixel-level features generated by the modified CLIP-based network are max-pooled for each mask, and the objective loss is computed with the text features.

We can notice that, regardless of whether the CLIP backbone is frozen, whether the CLIP network's architecture is modified, or whether additional network structures are appended to the side or rear of the CLIP network, the essence of these CLIP-based models lies in aligning image features with text features through contrastive learning.

The aforementioned methods hold a similar view with contrastive learning for point cloud pre-training. The difference is that the latter uses image-point cloud contrastive learning \citep{PPKT,SLidR} (some of the work uses data augmentation for single-modal contrastive learning). Most of them use SLIC \citep{slic}, SAM \citep{sam}, and SEEM \citep{SEEM} to guide mask segmentation and choose ResNet \citep{resnet} as the image encoder, which means that mask segmentation and mask feature generation are two completely independent modules. This not only increases the training time but also makes noise easily stack up across different models. At the same time, the lack of language guidance during segmentation will lead to a more random mask granularity. Fortunately, 2D open-vocabulary segmentation models perfectly address this issue, as we can not only extract labels and image embeddings from them but also obtain segmentation with appropriate granularity.

\begin{figure*}[tb]
  \centering
  \includegraphics[width=\linewidth]{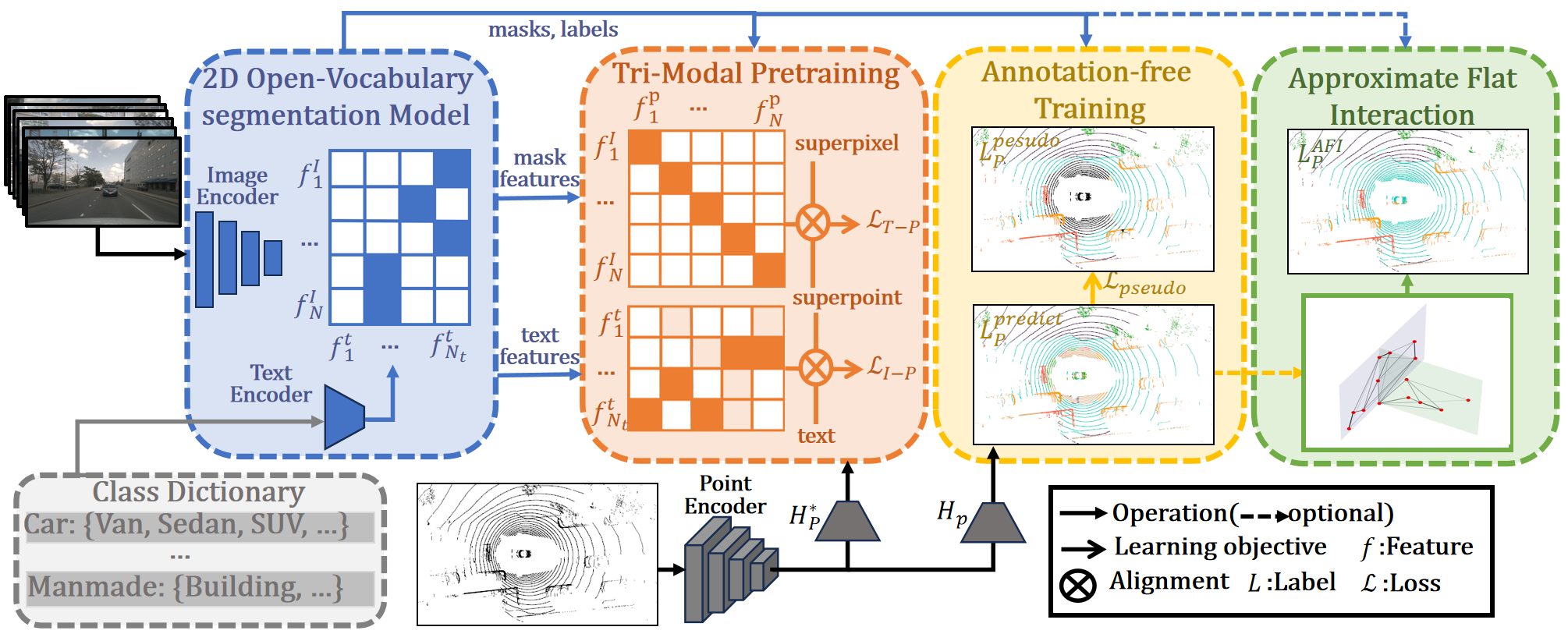}
  \caption{Overview of AFOV, which consists of two stages: Tri-Modal Pre-training (TMP) and Annotation-free training (AFOV-baseline). Both stages leverage masks and mask labels extracted from 2D open-vocabulary segmentation models, while mask features and text features are employed only in TMP. TMP enhances scene understanding through contrastive losses: superpixel-superpoint loss $\mathcal{L}_{I-P}$ and text-superpoint loss $\mathcal{L}_{T-P}$, while our baseline employs pseudo-labels to supervise the 3D network. Additionally, to bridge dataset classes and open vocabularies, we introduce a class dictionary. The Approximate Flat Interaction (AFI) optimizes the results by spatial structural analysis in a broad perception domain.}
  \label{fig:overview}
\end{figure*}

To summarize, we extract four interrelated, synchronously generated knowledge from CLIP-based open-vocabulary segmentation models: 1) images' segmentations as masks $M_\mathcal{I}$ from image set $\mathcal{I}$; 2) corresponding labels $L_{M}$ for $M_\mathcal{I}$; 3) image features corresponding to $M_\mathcal{I}$, denoted as $F_{M}$; and 4) text features $F_{T}$. Each of the above knowledge will play an important role in the following sections, as shown in Fig. \ref{fig:overview}.

\subsection{Baseline of AFOV}
\label{3.2}
% \subsubsection{Preliminaries}

Given a point cloud $\mathcal{P}=\{(p_n,e_n)|n=1,\ldots,N\}$, where $p_n\in \mathbb{R}^3$ represents the 3D coordinates of a point, $e_n\in \mathbb{R}^{E}$ denotes the point's features. ${L}=\{l_n|n=1,\ldots,N\}$ are the labels of $\mathcal{P}$ and $\mathcal{I}=\{i_k|k=1,\ldots,K\}$ represents the images captured by a synchronized camera at the same moment. In contrast to supervised methods, our task does not utilize labels ${L}$ during training. We choose to employ a simple way of generating pseudo-labels for point clouds with the assistance of image segmentation: With masks $M_\mathcal{I}=\{m_r|r=1,\ldots,R\}$ obtained from image set $\mathcal{I}$ as described in the above section, we use the labels ${L}_M$ corresponding to $M_\mathcal{I}$ as the pseudo-label ${L}_\text{pixel}^{\text{pseudo}}$ for pixels in every mask $m_r \in M_\mathcal{I}$. By leveraging known sensor calibration parameters, we establish a mapping $\Gamma_{\text{camera} \leftarrow \text{LiDAR}}$ to bridge the gap between domains of point clouds and images. Pseudo-labels ${L}_\mathcal{P}^{\text{pseudo}}=\{l_{n_0}^{\text{pseudo}}|n_0=1,\ldots,N_0\}$ for point clouds $\mathcal{P}$ are generated through ${L}_\text{pixel}^{\text{pseudo}}$ and mapping $\Gamma_{\text{camera} \leftarrow \text{LiDAR}}$. For a 3D backbone $\mathcal{F}_{\theta_p}:\mathbb{R}^{N\times(3+L)} \rightarrow \mathbb{R}^{N\times D}$ with the learnable parameter $\theta_p$, we train $\theta_p$ with pseudo-labels ${L}_\mathcal{P}^{\text{pseudo}}$. Given the sparsity of point clouds, it is obvious that $\Gamma_{\text{camera} \leftarrow \text{LiDAR}}$ is not surjective. It is important to note that $\Gamma_{\text{camera} \leftarrow \text{LiDAR}}$ is also not injective, as the projection area of LiDAR is not entirely covered by cameras, resulting in obvious pseudo-label-blank areas in point cloud $\mathcal{P}$. After knowledge distillation, these untrained regions exhibit label confusion, which will be discussed in the next section.

% \subsubsection{class dictionary}

To align open vocabularies with the stuff-classes of autonomous driving datasets, we employ a class dictionary $\mathcal{C}=\{{c_i}:[t_1^{c_i},\ldots,t_{n_{c_i}}^{c_i}]|i=1,\ldots,N^\mathcal{C}\}$, where $N^\mathcal{C}$ represents the number of stuff-classes.  Texts belonging to the same class $t_{j_1}, t_{j_2}\in c_i$ are uniformly mapped to the pseudo-label corresponding to $c_i$, which implies that points corresponding to $t_{j_1}$ and $t_{j_2}$ are positive samples for each other.

\subsection{Tri-Modal Contrastive Pre-training (TMP)}
\label{3.3}

In this section, we introduce Tri-Modal contrastive Pre-training (TMP). TMP innovatively integrates textual information into pre-training and removes the 2D backbone through pre-generation of the features, demonstrating excellent performance in both annotation-free training and fine-tuning. The illustration of TMP can refer to Fig. \ref{fig:overview}.

\subsubsection{Synchronous Generation of Knowledge}
To the best of our knowledge, most existing 3D pre-training methods \citep{PPKT,SLidR,seal} generate masks and mask features asynchronously for 2D-3D contrastive learning, which causes noise aggregation between different backbones. In TMP, we address this issue by synchronously generating masks and features before pre-training.

\subsubsection{Integrating Textual Information in TMP}
We aim to incorporate text-3D contrastive learning into pretraining. For different instances of the same category (e.g., Car A and Car B), they share the same textual features but have different image features. Theoretically, 2D-3D contrastive learning provides rich features from the instance level (needed for pretraining); text-3D contrastive learning, on the other hand, offers direct semantic embeddings for the 3D backbone from the semantic level (helpful for annotation-free segmentation). So the design intention of TMP is: Can integrating textual information into pre-training accomplish multiple tasks (pretraining and unannotated segmentation)? Experiment removing text-3D contrastive learning is conducted in Appendix B.

\begin{figure}[tb]
  \centering
  \includegraphics[width=0.9\linewidth]{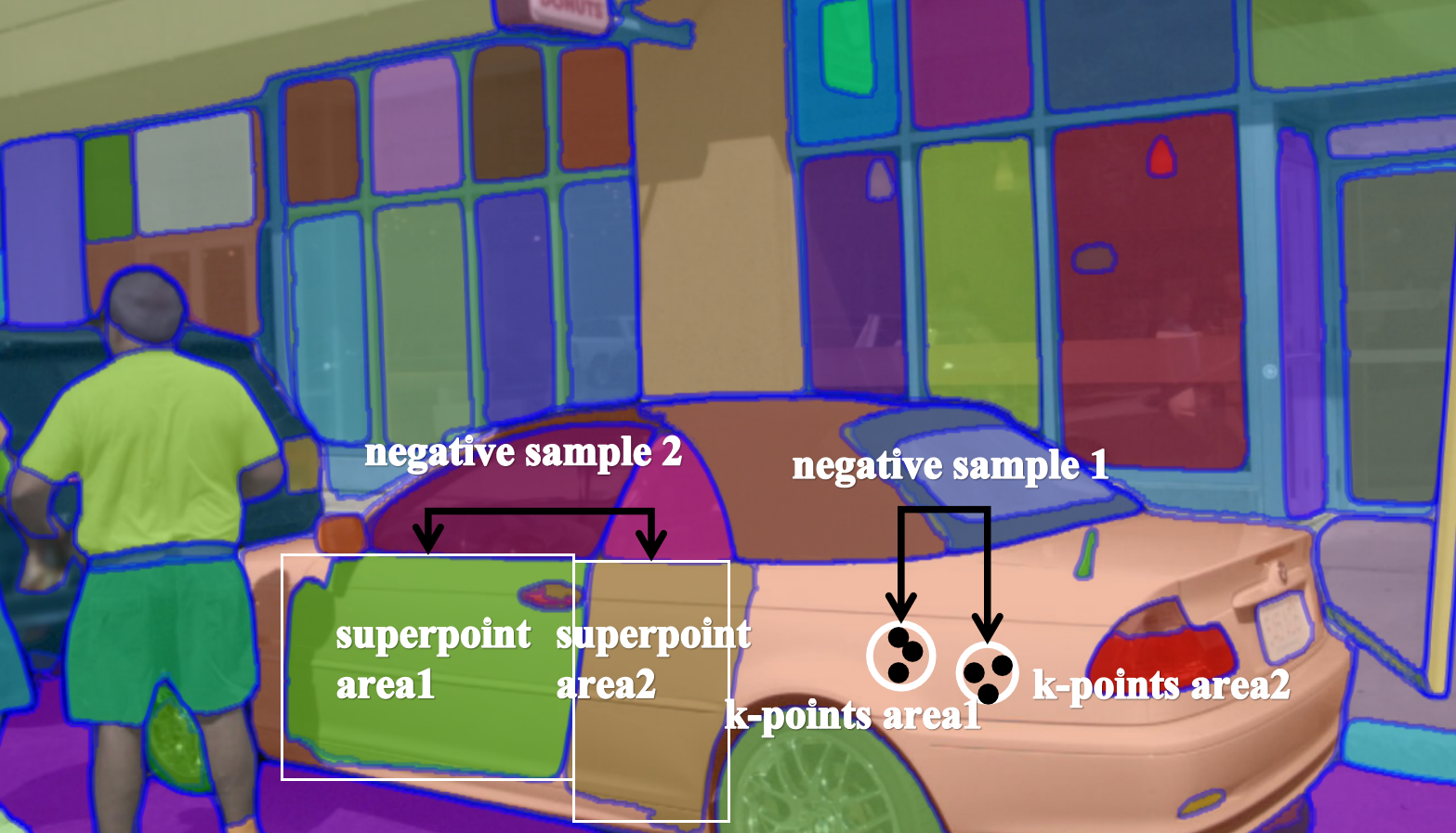}
  \caption{Illustrating two examples of potential "self-conflicts" based on SAM segmentation.}
  \label{fig:self_conflict}
\end{figure}

% \subsubsection{Superpixel-Superpoint Generation}
\subsubsection{"Self-Conflict"}
Regions with the same semantics may be divided into multiple parts (in 2D or 3D). During contrastive learning, different parts act as "negative samples" to each other, causing features of parts with the same semantics to be pushed apart in the feature space. This phenomenon is called "self-conflict". The usage of superpixels and superpoints for segmentation in SLidR \citep{SLidR} is inspiring, which partly alleviates the issue of "self-conflict" caused by point-level point set partitioning (\textit{e.g.} $k$-NN), such as "negative sample 1" in Fig. \ref{fig:self_conflict}. In Appendix B, we conduct an ablation study on the use of superpixels and superpoints, and the results demonstrate their effectiveness.

However, the "self-conflict" caused by the mask granularity randomness of some VLMs has not been properly addressed, such as "negative sample 2" in Fig. \ref{fig:self_conflict}. For instance, SAM \citep{sam} might separate the car-door and car-window because SAM is unaware of the appropriate segmentation granularity (both should belong to the same class: "car"). This issue arises due to the lack of textual assistance. When guided by the text, a 2D open-vocabulary segmentation model would treat the entire car as a whole, thereby avoiding "self-conflict". 

% On the contrary, 2D open-vocabulary segmentation models \citep{fcclip,san,CAT-Seg} perfectly fit the granularity by incorporating textual information defined in class dictionary $\mathcal{C}$. Also, as shown in Fig. \ref{fig:ov_segments}, the accuracy of the superpixels is greatly influenced by the models. Based on these observations, it is reasonable to generate superpixels using CLIP-based 2D open-vocabulary segmentation models.

\subsubsection{Superpixel-Superpoint Generation}

Given the point cloud $\mathcal{P}$ and images $\mathcal{I}$, we have generated masks $M_\mathcal{I}$ and use $\Gamma_{\text{camera} \leftarrow \text{LiDAR}}$ to map the labels $L_M$ of masks to $\mathcal{P}$. We regard the set of pixels with corresponding points in the same mask $m_r \in M_\mathcal{I}$ as a superpixel $S_r^{\text{pixel}} \in \mathcal{S}^{\text{pixel}}$, while the corresponding region of the point cloud as a superpoint $S_r^{\text{point}} \in \mathcal{S}^{\text{point}}$, $\mathcal{S}^{\text{pixel}}$ and $\mathcal{S}^{\text{point}}$ establish a bijection and ensuring $|\mathcal{S}^{\text{pixel}}|=|\mathcal{S}^{\text{point}}|=R^\mathcal{S}\leq |M_\mathcal{I}|=R$. Assuming the point cloud backbone $\mathcal{F}_{\theta_p}$ comes with an output head $H_p$, we replace $H_p$ with a trainable projection head $H_p^{*}$, projecting the point cloud feature $f_p$ of $p\in \mathcal{P}$ into a $D^{*}$-dimensional space such that $Dim_{f_M}=Dim_{f_T}=D^{*}$, here $f_M \in F_M$ and $f_T\in F_T$ refer to mask features and text features provided by the 2D open-vocabulary semantic segmentation models. Firstly, we apply average pooling and normalization to each group of pixel features ${F}_{r0}=\{f_p|p\in S_{r0}^{\text{point}}\}$ guided by superpoints to extract the superpoint embeddings $f_{r0}^{\text{p}}\in F^{\text{superpoint}}$. Then, we normalize $F_M$ as the superpixel embeddings $F^{\text{superpixel}}$ and consider the masks-corresponding text features as $F^{text}$. Finally, we employ a tri-modal contrastive loss to align $F^{\text{superpixel}}-F^{\text{superpoint}}$, $F^{\text{text}}-F^{\text{superpoint}}$.

\subsubsection{Tri-Modal Contrastive Loss}
Superpixel-guided contrastive learning operates at the object level or semantic level, rather than at the pixel or scene level. The contrastive loss between $F^{\text{superpixel}}$ and $F^{\text{superpoint}}$ is formulated as:

\begin{equation}
\begin{split}
    &\mathcal{L}_{I-P}=\mathcal{L}{(F_{\text {superpixel}}, F_{\text {superpoint }})} \\
    &=-\frac{1}{R'} \sum_{i=0}^{R'} \log \left[\frac{e^{(\langle f_{i}^{I}, f_{i}^{p}\rangle / \tau)}}{\sum_{j \neq i} e^{(\langle f_{i}^{I}, f_{j}^{p}\rangle / \tau)}+e^{(\langle f_{i}^{I}, f_{i}^{p}\rangle / \tau)}}\right],
\end{split}
\label{L-I-P}
\end{equation}
where $f_i^I\in F^{\text{superpixel}}$ is the feature of $i_{\text{th}}$ superpixel and $f_j^p\in F^{\text{superpoint}}$ is the feature of $j_{\text{th}}$ superpoint. $\langle \cdot \rangle$ denotes the cosine similarity and $\tau$ denotes the temperature coefficient. ${R'}$ is the mini-batch size.

Unlike the superpixel-superpoint contrastive loss, the text-superpoint contrastive loss does not exhibit "self-conflict" on classes of a dataset. However, to ensure the uniformity of knowledge in TMP, we retained the class dictionary $\mathcal{C}$ as discussed in \textbf{Baseline of AFOV}. In downstream tasks, texts of the same class $t_{j_1}, t_{j_2} \in c_i$ should be considered as positive samples for each other, so treating the point cloud regions corresponding to $t_{j_1}, t_{j_2}$ as "negative samples" will inadvertently cause "self-conflict". Therefore, for text $t_{j_0}\in c_i$, we utilize the text feature's cosine similarity $\langle f_{t_{j_0}},f_{t_{j_s}} \rangle$ weighted for other texts in the same class $\{t_{j_s}\in c_i, j_s=\{1,\ldots,n_{c_i}\}\neq j_0\}$ as "semi-positive" samples to compute $\mathcal{L}_{T-P}$:

\begin{equation}
\begin{split}
&\mathcal{L}_{T-P}=\mathcal{L}{(F_{\text {text }}, F_{\text {superpoint}})} \\
&=-\frac{1}{R'} \sum_{i=0}^{R'} \log \left[\frac{e^{(\langle f_{i}^{t}, f_{i}^{p}\rangle / \tau)}}{\sum_{t_j \neq t_i} e^{\left((1-\alpha_{i j})\langle f_{i}^{t}, f_{j}^{p}\rangle/ \tau\right)}+e^{(\langle f_{i}^{t}, f_{i}^{p}\rangle / \tau)}}\right],
\end{split}
\label{L-T-P1}
\end{equation}

\begin{equation}
\alpha_{i j}=\left\{\begin{array}{cc}
\left\langle f_{i}^{t}, f_{j}^{t}\right\rangle \quad  & t_{i}, t_{j} \text { in same class } \\
0 & \text { else }
\end{array}\right.
\label{L-T-P2}
\end{equation}
where $f_{i}^{t} \in F^{\text{text}}$ is the text feature corresponding to the mask. 

Tri-modal contrastive loss is calculated as:

\begin{equation}
\mathcal{L}_{TMP}=\alpha_{image}\mathcal{L}_{I-P}+\alpha_{text}\mathcal{L}_{T-P},
\label{L}
\end{equation}
$\alpha_{\text{image}}$, $\alpha_{\text{text}}$ are weights for $\mathcal{L}_{I-P}$ and $\mathcal{L}_{T-P}$.

% \subsubsection{Advantages}

TMP has the following advantages compared to previous pre-training methods: 1) TMP eliminates the need for image encodings during pre-training. It does not require an image backbone, which reduces the training time. 2) Addition of textual modality. Text-superpoint contrastive learning achieves semantic-level alignment, directly endowing the point cloud backbone with semantic features. 3) Synchronous generation of superpixels and $F^{\text{superpixel}}$. This not only controls segmentation granularity but also prevents the aggregation of noise between different backbones.

\subsection{Approximate Flat Interaction (AFI)}
\label{3.4}

Through the previous two sections of AFOV, we noticed three technical difficulties for annotation-free semantic segmentation that need to be solved: 1) Unprojected point cloud regions caused by differing or occluded fields of view (FoV) among devices. This directly results in the region of the point cloud outside the image FoV remains untrained for long periods, thus the untrained area suffers from serious label confusion. 2) Label noise in 3D. This arises from matching errors between cameras and LiDAR, as well as noise inherent in 2D open-vocabulary segmentation models. Therefore, we need a robust error correction mechanism for AFOV.

Inspired by Point-NN \citep{pointnn}, we propose a non-parametric network for Approximate Flat Interaction (AFI). AFI essentially expects points to interact only among points lying on approximate planes, thereby preserving labels of relatively small objects, \textit{e.g.}, pedestrians, and vehicles, within a broad perceptual domain. The process of AFI is formulated as:

\begin{equation}
L_{\mathcal{P}}^{AFI}=\operatorname{AFI}(L_{\mathcal{P}}^{predict},\mathcal{P},\gamma, (L_{\mathcal{P}}^{pseudo})),
\label{AFI}
\end{equation}
$L_{\mathcal{P}}^{predict}$ represents the predictions in \textbf{Baseline of AFOV}, and $\gamma$ indicates the minimum similarity between the directions when their directional features interact. $L_{\mathcal{P}}^{AFI}$, on the other hand, denotes the point cloud labels predicted by the function $AFI(\cdot)$. Meanwhile, we can choose to assist optimization through the pseudo-labels $L_{\mathcal{P}}^{pseudo}$ generated by 2D open-vocabulary segmentation models. A more detailed description of $AFI(\cdot)$ is stated in Appendix A.

During downsampling, AFI passes the directional features of the sampled center point through layer-wise interactions with neighboring points, and binds the correlation between two points based on 1) whether the two points are relevant in the same direction and 2) the tightness of the relevance between correlated directions. Through four rounds of downsampling, point-to-point interactions construct a network that, apart from points at the junctions, AFI ensures the surfaces formed by points on the same network approximate planes, thus tightly controlling interactions among points.

AFI is a robust error correction mechanism for AFOV. The advantages of AFI are evident. 1) Wide-sensing domain: The perception domain for the point clouds with AFI is wide and possesses strong spatial perception capabilities. 2) Detachability: The entire AFI is detachable, and the auxiliary module for 2D images within the AFI is detachable. The effectiveness of AFI can be referenced in Appendix B: Ablation Study.

\section{Experiments}
\subsubsection{Experiments Setup}
\subsubsection{Datasets}
\label{4.0.1}
To validate the performance of our model, multiple experiments on two large-scale autonomous driving datasets, nuScenes \citep{nuscenes} and SemanticKITTI \citep{semantickitti,KITTI} were conducted, as detailed in \textbf{Comparison Results} and \textbf{Ablation Study}. In nuScenes, there are 700 scenes for training, while the validation and test set each consist of 150 scenes, comprising a total of 16 semantic segmentation classes. During pre-training, only the train set was utilized, while we validated using specific scenes separated from the train set. SemanticKITTI has 19 classes, with its 22 sequences partitioned into specific train, validation, and test sets. 

\subsubsection{Implementation Details}
\label{4.0.2}
We followed the training paradigm of SLidR \citep{SLidR}, employed MinkowskiNet18 \citep{Minkowski-net} as the 3D backbone, and used a linear combination of the cross-entropy and the Lovász loss \citep{Lovász} as training objective in annotation-free and downstream tasks. For 2D open-vocabulary segmentation models, we employed FC-CLIP \citep{fcclip}, SAN \citep{san}, CAT-Seg \citep{CAT-Seg} for both TMP and annotation-free training, while using MaskCLIP \citep{MaskCLIP} as a control group. The generation of mask features and text features were synchronized with the masks and mask labels. FC-CLIP \citep{fcclip} employed panoptic segmentation, distinguishing different instances on thing-classes. MaskCLIP \citep{MaskCLIP}, SAN \citep{san}, and CAT-Seg \citep{CAT-Seg} utilized semantic segmentation, not distinguishing instances with the same semantics in both TMP and annotation-free training. Their mask features are selected as the average pool of pixel features in semantically identical regions. In Tri-Modal contrastive Pre-training (TMP), our network was pre-trained for 40 epochs on 4 V100 GPUs with a batch size of 4, which takes about 80 hours. 
% For annotation-free training in Sec. \ref{3.2} and other downstream tasks, the network was trained for 30 epochs on a single V100 GPU with a batch size of 16, which takes about 18 hours. 
For annotation-free training in \textbf{Baseline of AFOV} and other downstream tasks, the network was trained for 5 epochs and 30 epochs on a single V100 GPU, each task taking approximately 3 hours and a batch size of 16. On a 4090 GPU, annotation-free training for 5 epochs only took one hour.
The temperature coefficient $\tau$ in Eq. \ref{L-I-P},\ref{L-T-P1} was set to 0.07, and the optimal results achieved for Eq. \ref{L} when $\alpha_{\text{image}}=\alpha_{\text{text}}=0.5$. In Eq. \ref{AFI}, the minimum similarity $\gamma$ between the directions when their directional features interact, was set to 0.995, implying that the maximum angular disparity of two interact point features is about 5.7°. In the network structure of AFI, downsampling was performed four times, with the downsampling rate being $1/3$ for the last three times. Additional details about AFI are provided in Appendix A.

\begin{table}[tb]
\caption{Comparisons of 3D annotation-free semantic segmentation results (\% mIoU) on nuScenes \citep{nuscenes} \emph{val} set.}%title
\centering
\scalebox{0.8}
{\begin{tabular}{c|ccc|c}% four columns
\bottomrule
% \begin{tabular}{@{}c@{}}use label\\proportion\end{tabular}
Method & 
\begin{tabular}{@{}c@{}}Annotation\\Ratio\end{tabular} & 
\begin{tabular}{@{}c@{}}Image\\Infer\end{tabular} & 
\begin{tabular}{@{}c@{}}3D\\backbone\end{tabular} & mIoU \\
\midrule
CLIP2Scene [CVPR'23] & 0\% & X & SPVCNN & 20.80 \\
Chen et al. [NeurIPS'23] & 0\% & X & MinkowskiNet & 26.80 \\
% OpenScene3D+LSeg[CVPR'23] & 0\% &  & MinkowskiNet & 33.78  \\
OpenScene [CVPR'23] & 0\% & X & MinkowskiNet & 41.30 \\
OpenScene-LSeg & 0\% & X & MinkowskiNet & 35.50 \\
OV3D [CVPR'24] & 0\% & X & MinkowskiNet & 44.60 \\
% AFOV(ours)+Mask-clip & 0\% & X & MinkowskiNet & 30.35 \\
% AFOV(ours)+FC-CLIP & 0\% & X & MinkowskiNet & 43.28 \\
% AFOV(ours)+CAT-Seg & 0\% & X & MinkowskiNet & 42.83 \\
AFOV(ours)+SAN & 0\% & X & MinkowskiNet & \textbf{47.73} \\
\midrule
% OpenScene+LSeg[CVPR'23] & 0\% & \checkmark & MinkowskiNet & 34.90 \\
OpenScene [CVPR'23] & 0\% & \checkmark & MinkowskiNet & 36.30 \\
OpenScene-LSeg & 0\% & \checkmark & MinkowskiNet & 42.10 \\
% AFOV(ours)+Mask-clip \citep{MaskCLIP} & 0\% & \checkmark & MinkowskiNet & 27.76 \\
% AFOV(ours)+FC-CLIP & 0\% & \checkmark & MinkowskiNet & 43.64 \\
% AFOV(ours)+CAT-Seg & 0\% & \checkmark & MinkowskiNet & 43.64 \\
AFOV(ours)+SAN & 0\% & \checkmark & MinkowskiNet & \textbf{47.89} \\
\midrule
- & 100\% &  & MinkowskiNet & 74.66 \\
% - & 100\% &  & MinkowskiNet34 & - \\
% - & 100\% &  & SPVCNN[] & - \\
\bottomrule
\end{tabular}}
\label{tab:anno-free Contrastive}
\end{table}

\begin{figure*}[tb]
  \centering
  \includegraphics[width=0.9\linewidth]{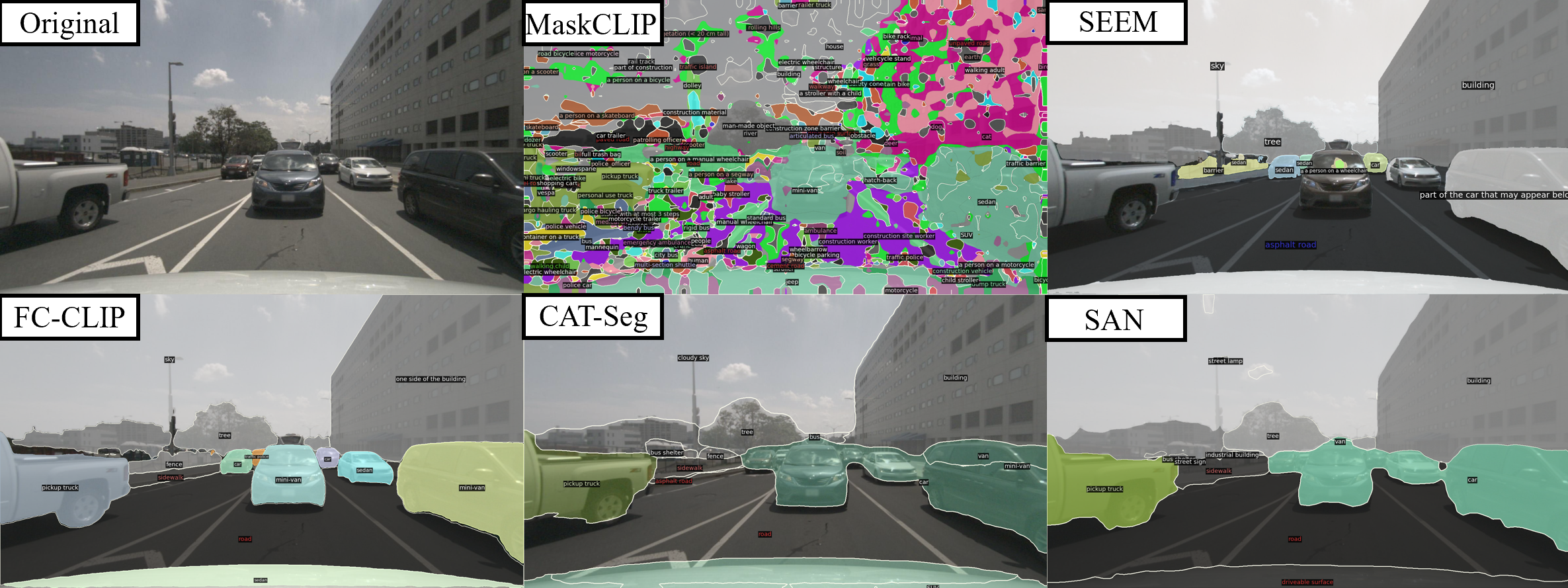}
  \caption{Illustration of image segmentation results of various 2D open-vocabulary segmentation models. We observe that MaskCLIP (pixel-level CLIP) exhibits label confusion and high error rates in semantic segmentation. The output of SEEM not only suffers from missing masks but also contains incorrect mask annotations. More results are provided in Appendix D.}
  \label{fig:ov_segments}
\end{figure*}

\subsection{Comparison Results}
\label{4.1}
\subsubsection{Annotation-free Semantic Segmentation} 
In Tab. \ref{tab:anno-free Contrastive}, we compare AFOV with the most closely related works on 3D semantic segmentation using the unannotated data of nuScenes: CLIP2Scene \citep{CLIP2Scene} designs a semantic-driven cross-modal contrastive learning framework; Chen \textit{et al.} \citep{Towards_label-free} utilizes CLIP and SAM for denoising; OpenScene \citep{OpenScene} extracts 3D dense features from an open-vocabulary embedding space using multi-view fusion and 3D convolution; OV3D \citep{OV3D} seamlessly aligning 3D point features with entity text features. The optimal result of AFOV's single-modal annotation-free segmentation reaches 47.73\% mIoU, surpassing the previous best method by 3.13\% mIoU. Under image assistance, it achieves 47.89\% mIoU. The gap between AFOV and the fully supervised same backbone is only -26.93\% mIoU.

Compared to the multi-task capability of VLMs, state-of-the-art 2D open-vocabulary segmentation models demonstrate greater capability in specialized domains, as shown in the Fig. \ref{fig:ov_segments}. Selecting professional teacher models enhances the performance of student models effectively.

\begin{table}[tb]
\caption{Comparisons (\% mIoU) of different pre-training methods pre-trained on nuScenes \citep{nuscenes} and fine-tuned on nuScenes and SemanticKITTI \citep{semantickitti}. LP denotes linear probing with frozen backbones.}%title
\centering
\scalebox{0.78}{
\begin{tabular}{c|ccc}% four columns
\toprule
% \begin{tabular}{@{}c@{}}use label\\proportion\end{tabular}
\multirow{2}{*}{3D Initialization} & \multicolumn{2}{c}{nuScenes} & KITTI \\
 & 100\%LP & 1\%Fine-tuning & 1\%Fine-tuning \\
\midrule
Random & 8.10 & 30.30 & 39.50 \\
PointConstrast [ECCV'20] & 21.90 & 32.50 & 41.10 \\
DepthConstrast [ICCV'21]  & 22.10 & 31.70 & 41.50 \\
PPKT [arXiv'21] & 35.90 & 37.80 & 44.00 \\
SLidR [CVPR'22] & 38.80 & 38.30 & 44.60 \\
ST-SLidR [CVPR'23] & 40.48 & 40.75 & 44.72 \\
Seal [NeurIPS'23] & 44.95 & 45.84 & 46.63 \\
VLM2Scene [AAAI'24] & 51.54 & 47.59 & 47.37 \\
ScaLR [CVPR'24] & 42.40 & 40.50 & - \\
AFOV-TMP(ours)+FC-CLIP & 44.24 & 45.73 & 47.02 \\
AFOV-TMP(ours)+CAT-Seg & 43.95 & 46.61 & \textbf{48.14} \\
AFOV-TMP(ours)+SAN & 46.29 & 47.60 & 47.72 \\
AFOV(ours)+FC-CLIP & 52.92 & 50.58 & 45.86 \\
AFOV(ours)+CAT-Seg & 51.02 & 49.14 & 47.59 \\
AFOV(ours)+SAN & \textbf{56.35} & \textbf{51.75} & 46.60 \\
\bottomrule
\end{tabular}}
\label{tab:downstream Contrastive}
\end{table}

\subsubsection{Comparisons among 3D Pre-training Methods} 
We compared the performance of AFOV-TMP (only employing TMP) and AFOV (employing both steps) against other state-of-the-art methods on multiple downstream tasks in nuScenes and SemanticKITTI (all pre-trained on nuScenes), as shown in Tab. \ref{tab:downstream Contrastive}. All methods utilized MinkowskiNet as the 3D backbone. Most of the compared state-of-the-art methods utilize point cloud-image contrastive learning. SLidR \citep{SLidR} and ST-SLidR \citep{ST-SLidR} employ superpoint-superpixel correspondence granularity; ScaLR \citep{scalr} scales the 2D and 3D backbones and pretraining on diverse datasets; while SEAL \citep{seal}, similar to VLM2Scenes \citep{vlm2scene}, employs VLMs in distilling. Our approach achieved optimal results with 1\% data fine-tuning on nuScenes and SemanticKITTI, reaching 51.75\% mIoU and 48.14\% mIoU, respectively, demonstrating a respective improvement of +21.45\% mIoU and +8.64\% mIoU versus random initialization. Compared to the previously best results, AFOV exhibited enhancements of +4.16\% mIoU and +0.77\% mIoU, respectively. Remarkably, the results of the fully supervised linear probing task on nuScenes reached 56.35\% mIoU, displaying an improvement of +4.81\% mIoU.

\subsection{Ablation Study}
\label{4.2}
We conducted a series of ablation experiments on nuScenes. The ablation targets included different teacher models, TMP, AFI, \textit{etc}. The results obtained validate the effectiveness of our designs, especially TMP and AFI. Please refer to Appendix B for details of the ablation study.

\section{Conclusion}
We propose AFOV, a versatile two-stage annotation-free framework that serves for both 3D pre-training and annotation-free semantic segmentation, achieving state-of-the-art performance across multiple experiments. The key to AFOV is to leverage the high-quality knowledge of 2D open-vocabulary segmentation models. Moreover, We propose Tri-Modal contrastive Pre-training (TMP) and Approximate Flat Interaction (AFI) for the first time.

% We hope that this work will contribute to more in-depth research on 2D-3D transfer learning in the future and expect the emergence of annotation-free object detection (multi-task) models.

% As the trend towards data-driven 3D approaches becomes increasingly evident, future research will likely focus more on efficiently generating or utilizing data.  
We hope that our work will contribute to more in-depth research on 2D-3D transfer learning. Additionally, to the best of our knowledge, there is currently a lack of work on annotation-free training in other 3D perception tasks such as object detection, trajectory tracking, occupancy grid prediction. We expect the emergence of other annotation-free 3D perception approaches.

%%%%%%%%%%%%%%%%%%%%%%%%%%%%%%%%%%%%%%%%%%%%%%%%%%%%%%

\section{Acknowledgments}
This work is supported by the Beijing Natural Science Foundation (L245025) and the Joint Development of Multi-modal Parallel LiDARs with Waytous Inc.

% \bigskip
% \noindent Thank you for reading these instructions carefully. We look forward to receiving your electronic files!

\bibliography{aaai25}
\clearpage

\appendix
\section{A: Details of AFI}
\label{Appendix_A:AFI}

\begin{figure*}[tb]
  \centering
  \includegraphics[width=\linewidth]{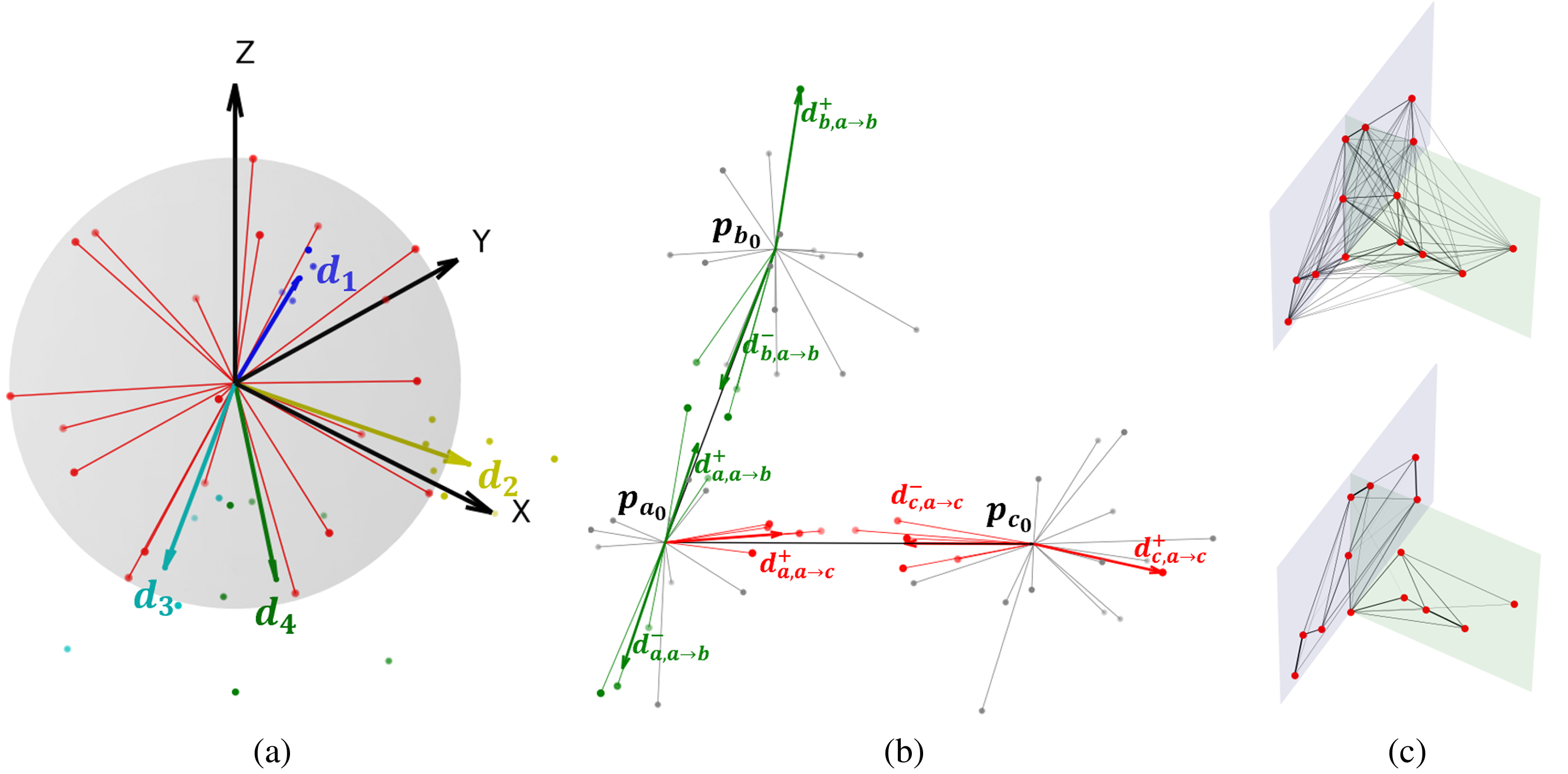}
  \caption{Illustration of AFI: (a) Demonstration of Fibonacci lattice. (b) Approximate-parallel interactions between points. (c) The distinction between distance-weighted interaction (above) and AFI (below).}
  \label{fig:AFI}
\end{figure*}

Firstly, we remove the predictions in the label-confused area. Then, we utilize pseudo-labels ${L}_\mathcal{P}^{\text{pseudo}}$ generated in \textbf{Baseline of AFOV} to randomly cover model-predicted labels (note that the coverage step is optional). The probability $P(d_p)$ for random covering point $p$ is defined as follows:

\begin{equation}
\begin{aligned}
P(d_p) &= \frac{\beta \cdot e^{-d_p/T}}{1 + \beta \cdot e^{-d_p/T}}, & \quad T &= \frac{S}{\ln \beta}.
\end{aligned}
\label{equ:afi_1}
\end{equation}

Here, variable $d_p$ represents the horizontal distance of point $p$, $\beta$ donates the ratio of the occurrence probabilities of pseudo labels and predicts labels when $d_p=0$, and $S$ represents the horizontal distance when $P(S)=0.5$.
We set $\beta=4$ and $S=15$ in Eq. \ref{equ:afi_1}.

Secondly, we need to fill large unannotated regions and eliminate noise introduced by 2D open-vocabulary segmentation models:

For a set of regional point clouds $\mathcal{P}'$ with corresponding features $F'$, assuming the center of $\mathcal{P}'$ is $p_0$, We try to perform directional clustering of relative coordinates for points $\mathcal{P}'$ excluding $p_0$. Considering the limited temporal effectiveness of conventional clustering methods, a novel way is explored:

We obtain the spherical Fibonacci lattice $FL=\{fl_i|i=0,\ldots,M-1\}$ of the sphere by the following equation, where for $\forall i\in0,\ldots,M-1$, the normal vector for each point on the grid is denoted as:
\begin{equation}
    \left.fl_{i}=\left\{\begin{array}{l}{{z_{i}=(2i-1)/M-1}}\\{{x_{i}=\sqrt{1-z_{i}^{2}}\cdot\cos(2\pi i\phi)}}\\{{y_{i}=\sqrt{1-z_{i}^{2}}\cdot\sin(2\pi i\phi)}}\\{\phi=\frac{\sqrt{5}-1}{2}}\\\end{array}\right.\right.,
\label{equ:afi_2}
\end{equation}
where $\phi$ is the golden ratio. Normal vectors $FL$ is shown in Fig. \ref{fig:AFI} (a).

After that, we perform clustering $\{p_n|C_n=i\}$ by calculating the cosine similarity between the relative coordinates of $p_n$ to $p_0$ and normal vectors $FL$.
\begin{equation}
\begin{aligned}
    C_{n}&=\operatorname*{argmax}_{i = 0,\ldots,M-1}\left(\langle(p_{n}-p_{0}), fl_{i} \rangle\right) \\
    &= \operatorname*{argmax}_{i = 0,\ldots,M-1}\left(\frac{(p_{n}-p_{0})\cdot fl_{i}}{|p_{n}-p_{0}|}\right)
\end{aligned}
\label{equ:afi_3}
\end{equation}

After clustering, the mean direction vector ${d_i}\in {D}^{\prime}$, direction feature ${f_i}\in {F}^{\prime}$, and correlation ${l_i}\in {L}^{\prime}$ are computed for each normal vector $fl_i$, where $i=0,\ldots,M-1$:
\begin{equation}
    d_{i}=\operatorname{avg}_{C_{n}=i}(p_{n}-p_{0}), \quad f_{i}=\sum_{C_{n}=i}f_{n}, \quad l_{i}=\sum_{C_{n}=i}l_{n},
\label{equ:afi_4}
\end{equation}
where $\operatorname{avg}(\cdot)$ denotes the average function, and $l_n$ represents the correlation between each point $p_n$ and the central point $p_0$. It is worth noting that during the first downsampling, the correlation $l_{i}$ is the proportion of points belonging to the same direction, \textit{ie.}, $l_n=1/N$. However, in subsequent downsampling steps, the correlation $l_n$ is obtained through the interaction among point cloud $\mathcal{P}'$ and other neighbor point clouds. This implies that we need to transmit the correlations $L'$ and the mean direction vectors $D'$ along with the coordinates and features from the previous layer.

As shown in Fig. \ref{fig:AFI} (b), in downsamplings, assuming the mean direction vectors and correlations for two interacting point clouds $\mathcal{P}_a$, $\mathcal{P}_b$ are ({$D_a$, $L_a$}) and ({$D_b$, $L_b$}), respectively. Assuming the direction from the center $p_{a0}$ of point cloud $\mathcal{P}_a$ to the center $p_{b0}$ of point cloud $\mathcal{P}_b$ is the positive direction, we calculate the cosine similarity $S_{\cdot,\cdot}^{a\rightarrow b}$ between the vector connecting two central points and the mean direction vectors $d_{i}\in D_a$, $d_{j}\in D_b$:
\begin{equation}
    S_{a,i}^{a\rightarrow b}=\langle (p_{a_{0}}-p_{b_{0}}),d_i \rangle ,for\ d_{i}\in D_{a}
\label{equ:afi_5}
\end{equation}
\begin{equation}
    S_{b,j}^{a\rightarrow b}=\langle (p_{a_{0}}-p_{b_{0}}),d_j \rangle ,for\ d_{j}\in D_{b}
\label{equ:afi_6}
\end{equation}

Afterward we obtain the index sets of approximate-parallel vectors in positive and negative directions:
\begin{equation}
    \begin{array}{l}
    I_{a}^{+}=\left\{i \mid S_{a, i}^{a\rightarrow b}>\gamma\right\} ,\hspace{0.5cm} d_{a,a\rightarrow b}^{+}=\max _{i \in I_{a}^{+}}\left(\left|d_{i}\right|\right),\\
    I_{a}^{-}=\left\{i \mid S_{a, i}^{a \rightarrow b}<-\gamma\right\} ,\hspace{0.5cm} d_{a,a\rightarrow b}^{-}=\max _{i \in I_{a}^{-}}\left(\left|d_{i}\right|\right),\\
    I_{b}^{+}=\left\{j \mid S_{b, j}^{a\rightarrow b}>\gamma\right\} ,\hspace{0.5cm}d_{b,a\rightarrow b}^{+}=\max _{i \in I_{b}^{+}}\left(\left|d_{i}\right|\right),\\
    I_{b}^{-}=\left\{j \mid S_{b, j}^{a\rightarrow b}<-\gamma\right\},\hspace{0.5cm}d_{b,a\rightarrow b}^{-}=\max _{i \in I_{b}^{-}}\left(\left|d_{i}\right|\right),
    \end{array}
\label{equ:afi_7}
\end{equation}
where $\gamma=0.995$ represents the minimum arccosine value of two vectors recognized as approximately parallel, which is the same as defined in Eq. \ref{AFI}. $d_{a,a\rightarrow b}^{+}$ represents the distance of $I_{a}^{+}$, and so forth.

Through the aforementioned sets of approximate-parallel vectors, we can obtain the correlation $l_{a, b}^{*}$ from point cloud $\mathcal{P}_b$ to $\mathcal{P}_a$, which is computed by multiplying two components: the correlation along the approximately parallel direction $l_{a, b}^{c,*}$, and the correlation based on the distance $l_{a, b}^{d,*}$:
\begin{equation}
l_{a, b}^{c,{*}}=\left(\sum_{\substack{i \in I_{a}^{+} \\
\text {or } i \in I_{a}^{-}}} l_{i}\right) \times\left(\sum_{\substack{j \in I_{b}^{+} \\
\text {or } j \in I_{b}^{-}}} l_{j}\right)
\label{equ:afi_8}
\end{equation}
\begin{equation}
    l_{a, b}^{d, *}=\frac{(d_{a,a\rightarrow b}^{+}+d_{a,a\rightarrow b}^{-}) \times(d_{b,a\rightarrow b}^{+}+d_{b,a\rightarrow b}^{-})}{(\left|p_{a_{0}}-p_{b_{0}}\right|+d_{a,a\rightarrow b}^{-}+d_{b,a\rightarrow b}^{+})^{2}}
\label{equ:afi_9}
\end{equation}

% \begin{equation}
%     \text { where }|d_{a}^{+}=\max _{i \in I_{a}^{+}}\left(\left|d_{i}\right|\right) \text { Others by analogy }
% \end{equation}

\begin{equation}
    l_{a, b}^{*}=l_{a, b}^{c, *}\times l_{a, b}^{d, *}
\label{equ:afi_10}
\end{equation}

During the last three downsamplings, for point cloud $\mathcal{P}'=\{p_n|n=1,\ldots,N\}$, we obtain the correlation $l_{n_0,k}^{*}$ of every surrounding point $p_k \in \mathcal{P}_{K_{n_0}}$ around a central point $p_{n_0}$ through Eq. \ref{equ:afi_5},\ref{equ:afi_6},\ref{equ:afi_7},\ref{equ:afi_8},\ref{equ:afi_9},\ref{equ:afi_10}. Among them, $\mathcal{P}_{n_0}:(D_{n_0}, L_{n_0})$ and $\mathcal{P}_k:(D_k, L_k)$ utilized in Eq. \ref{equ:afi_5},\ref{equ:afi_6} correspond to the computation results of point $p_{n_0}$ and $p_k$ in the previous layer. Afterwards, $l_{n_0,k}^{*}$ is converted to a probability distribution $l_{n_0,k}$ using softmax:
\begin{equation}
    \{l_{{n_0}, k}\}=\operatorname{softmax}_{k=1,..., K_{n_0}}\left(\{l_{{n_0}, k}^{*}\}\right)
\label{equ:afi_11}
\end{equation}

Through Eq. \ref{equ:afi_5},\ref{equ:afi_6},\ref{equ:afi_7},\ref{equ:afi_8},\ref{equ:afi_9},\ref{equ:afi_10},\ref{equ:afi_11}, we use the mean direction vectors $\{D_n|n=1,\ldots, N\}$, correlations $\{L_n|n=1,\ldots, N\}$ and features $\{f_{p_n}|n=1,\ldots, N\}$ generated before each downsampling step to assist in generating the new correlation $l_{n_0,k}$ for each point $p_k$ around $p_{n_0}$. Then we use Eq. \ref{equ:afi_4} to obtain $L_{n_0}'=\{l_i^{n_0}|i=0,\ldots,M-1\}$ from $\{l_{n_0,k}|p_k \in \mathcal{P}_{K_{n_0}}\}$ and $F_{n_0}'=\{f_i^{n_0}|i=0,\ldots,M-1\}$ from $\{f_{p_k}\}$. Afterwords $L_{n_0}'$ acts as weights for the directional features $F_{n_0}'$ and contributes to the generation of central point features (Eq. \ref{equ:afi_12}). Throughout the entire process, we bind the correlation between two points to 1) whether they are relevant in the same direction (corresponding to $l_{a, b}^{c,*}$ in Eq. \ref{equ:afi_8}), and 2) how close the relevance is between the correlated directions (corresponding to $l_{a, b}^{d,*}$ in Eq. \ref{equ:afi_9}). Through four rounds of downsampling, point-to-point interactions construct an approximate flat network, thus tightly controlling feature interactions among points (Fig. \ref{fig:AFI} (c)).

\subsection{Neural Network Architecture}

\textbf{Encoder:} For point cloud $\mathcal{P}$, we use one-hot encoding of the predict labels $L_{\mathcal{P}}^{predict}$ in Eq. \ref{AFI} as pseudo-features $f_{p_n}$ for point $p_n$. Within each layer of the encoder, we first employ the farthest point sampling (FPS) followed by $K_{n_0}$-nearest neighbors ($k$-NN) for downsampling. It is necessary to transmit the mean direction vectors ${D_n}$ and correlations ${L_n}$ from the previous layer's output during downsampling. Then, using Eq. \ref{equ:afi_2},\ref{equ:afi_3},\ref{equ:afi_4},\ref{equ:afi_5},\ref{equ:afi_6},\ref{equ:afi_7},\ref{equ:afi_8},\ref{equ:afi_9},\ref{equ:afi_10},\ref{equ:afi_11}, we calculate the mean direction vectors $D_n^{\prime}=\{d_i^n|i=0,\ldots,M-1\}$, features ${F_n}^{\prime}=\{f_i^n|i=0,\ldots,M-1\}$ and correlations $L_n^{\prime}=\{l_i^n|i=0,\ldots,M-1\}$ for each point $p_n$. Next, for each central point $p_{n_0}$, we use the following formula to generate the new point feature $f_{p_{n_0}}^{\prime}$:

% \vspace*{0.6cm}

\begin{equation}
\label{equ:afi_12}
\begin{aligned}
f_{p_{n_0}}^{\prime} = \operatorname{softmax}&\left(\sum_{i=0}^{M-1} \left(l_{i}^{n_0} \times f_{i}^{n_0} + 0.1 \times f_{p_{n_0}}^{*} \right.\right. \\
&\left. + 1 \times 10^{-8} \times \operatorname*{Maxpooling}_{k \in K_{n_0}}\left(f_{p_k}^{*}\right) \Big)\right),
\end{aligned}
\end{equation} 
where $f_{p_{n_0}}^{*}$ is the original point feature, and $\{f_{p_k}^{*}\}$ represents original features of points in the neighborhood. The softmax operates on the features.

\textbf{Decoder:} Similar to the operations in the encoder, the upsampling layer of the decoder utilizes the directional correlations $l_i^{n_0}\in L_{n_0}^{\prime}$ generated at each layer of the encoder. After selecting $K_{n_0}'$ nearest neighbors, the impact of distance on the upsampling weights is discarded:

\begin{equation}
\begin{aligned}
    f_{p_{n_0}}^{\prime \prime}=\operatorname{softmax}\left(\sum_{k \in K_{n_0}'} \operatorname*{softmax}_{k\in K_{n_0}'}\left(\sum_{\left|S_{n_0, i}^{{n_0} \rightarrow k}\right|>\gamma}^{{i \in 0,\ldots,M-1}} l_i^{n_0,k}\right) f_{p_k}^{**} \right.\\
    + f_{p_{n_0}}'\Bigg),
\end{aligned}
\end{equation}
where $f_{p_k}^{**}$ represents the original features of points in the neighborhood. $l_i^{n_0,k}$ denotes the correlation between $p_{n_0}$ and neighbor $p_k$ in direction $i \in \{0,\ldots,M-1\}$, $f_{{p}_{n_0}}'$ represents the features of corresponding points in the encoder's same layer (Eq. \ref{equ:afi_12}). The definition of $S_{n_0, i}^{n_0 \rightarrow k}$ is consistent with Eq. \ref{equ:afi_5},\ref{equ:afi_6}.

% Finally, we obtain the labels $L_{\mathcal{P}}^{AFI}=\operatorname{argmax}(f_{p_{n}}^{\prime \prime})$ at the final layer of the decoder.

\begin{table*}[tb]
\caption{Ablation study on different teacher models of AFOV's annotation-free semantic segmentation(\% mIoU) on nuScenes \citep{nuscenes}.}%title
\centering
\scalebox{0.81}
{\begin{tabular}{c|cccccc}% four columns
\toprule
% \begin{tabular}{@{}c@{}}use label\\proportion\end{tabular}
\begin{tabular}{@{}c@{}}2D Teacher\\Model\end{tabular} & Baseline & \hspace{0.2cm}+TMP\hspace{0.1cm} & \hspace{0.2cm} +AFI\hspace{0.4cm} 
& \begin{tabular}{@{}c@{}}+Image\\Inference\hspace{0.1cm}\end{tabular}
& \begin{tabular}{@{}c@{}}+TMP\\+AFI\end{tabular} 
& \begin{tabular}{@{}c@{}}+TMP+AFI\\+Image Inference\end{tabular} \\
\midrule
MaskCLIP(ECCV'22) \citep{MaskCLIP} & 25.49 & - & 30.35 & 27.76 & - & - \\
FC-CLIP(NeurIPS'23) \citep{fcclip} & 39.00 & 41.70 & 41.89 & 42.44 & 43.28 & 43.64 \\
CAT-Seg(CVPR'24) \citep{CAT-Seg} & 38.45 & 40.92 & 41.79 & 42.50 & 42.83 & 43.64 \\
SAN(CVPR'23) \citep{san} & 44.16 & 47.42 & 46.26 & 46.92 & 47.73 & 47.89 \\
\midrule
\end{tabular}}
\label{tab:anno-free ablation}
\end{table*}

\begin{table*}[tb]
\caption{Open-vocabulary semantic segmentation performance on COCO \citep{coco} and pseudo-label accuracy(Acc) of different 2D open-vocabulary segmentation models.}%title
\centering
\scalebox{0.81}
{\begin{tabular}{c|cccc|ccc}% four columns
\toprule
% \begin{tabular}{@{}c@{}}use label\\proportion\end{tabular}
\multirow{2}{*}{2D Teacher Model}& \multicolumn{4}{c} {COCO (\% mIoU)} \vline 
  &  \multirow{2}{*}{Segment task}
  & pseudo-label
  & pseudo-label  \\
  \hspace{1cm}& A-847 & PC-459 & A-150 & PC-59
  &&coverage rate
  & Acc  \\
\midrule
MaskCLIP(ECCV'22) \citep{MaskCLIP} &8.2 &10.0 &23.7 &45.9 & Semantic & 55.00 & 50.18  \\
FC-CLIP(NeurIPS'23) \citep{fcclip} &14.8 &18.2 &34.1 &58.4 & panoptic  & 45.95 & 78.96\\
CAT-Seg(CVPR'24) \citep{CAT-Seg} & 10.8& 20.4&31.5 &62.0 & Semantic & 55.00& 77.05\\
SAN(CVPR'23) \citep{san} &13.7 &17.1 &33.3 &60.2 & Semantic & 55.00 & 80.38 \\
\bottomrule
\end{tabular}}
\label{tab:ACC}
\end{table*}

\begin{table}[tb]
\caption{Ablation study of different targets based on CAT-Seg \citep{CAT-Seg}.  $M_I$ and $F_M$ respectively denote whether employing segmentation results or mask features from CAT-Seg. If $F_M$ is not utilized, ResNet50 \citep{resnet} is integrated as the image backbone. $\mathcal{L}_{T-P}$ indicates the adoption of text-superpoint contrastive loss. $SP$ means whether to use superpixel and superpoint. $\mathcal{C}$ represents class dictionary, while $+2D$ indicates the incorporation of image inference in AFI.}%title
\centering
\scalebox{0.81}
{\setlength{\tabcolsep}{3pt}
\begin{tabular}{ccccccc|cc}% four columns
\toprule
& TMP & & &\multirow{2}{*}{$\mathcal{C}$} & \multirow{2}{*}{AFI}
& \multirow{2}{*}{$+2D$} & \multirow{2}{*}{Annotation-free} &\multirow{2}{*}{1\% Fine-tune}\\
$M_I$ & $F_M$ & $\mathcal{L}_{T-P}$ & $SP$ & & & & &\\

\midrule
- & - & - & - &\checkmark &  &  & \begin{tabular}{@{}c@{}}AFOV-baseline\\38.45\end{tabular} & \begin{tabular}{@{}c@{}}SLidR\\38.30\end{tabular} \\
\midrule
\checkmark &  &  & \checkmark & \checkmark &  &  & 39.62\textcolor{red}{(+1.17)} & 43.31\textcolor{red}{(+5.01)} \\
\checkmark &  & \checkmark & \checkmark & \checkmark &  &  & 40.49\textcolor{red}{(+2.04)} & 45.83\textcolor{red}{(+7.53)} \\
\checkmark & \checkmark &  & \checkmark & \checkmark &  &  & 39.84\textcolor{red}{(+1.39)} & 43.70\textcolor{red}{(+5.40)} \\
\checkmark & \checkmark & \checkmark & & \checkmark &  &  & 39.34\textcolor{red}{(+0.89)} & 41.87\textcolor{red}{(+3.57)} \\
\checkmark & \checkmark & \checkmark & \checkmark & \checkmark &  &  & 40.92\textcolor{red}{(+2.47)} & 46.61\textcolor{red}{(+8.31)} \\
\midrule
 - & - & - & - &  &  &  & 32.31\textcolor{green}{(-6.14)} & - \\
\checkmark & \checkmark & \checkmark & \checkmark & \checkmark &  &  & 40.92\textcolor{red}{(+2.47)} & - \\
\checkmark & \checkmark & \checkmark & \checkmark & \checkmark & \checkmark &  & 42.83\textcolor{red}{(+4.38)} & - \\
\checkmark & \checkmark & \checkmark & \checkmark & \checkmark & \checkmark & \checkmark & 43.64\textcolor{red}{(+5.19)} & - \\
\bottomrule
\end{tabular}}
\label{tab:catseg ablation}
\end{table}

\begin{table}[tb]
\caption{Ablation Study on 2D Feature Pre-generation. Experiments were conducted based on whether the mask features were pre-generated during CAT-Seg pre-training (if the features were not pre-generated, the 2D backbone network in TMP exists and all layers except the feature output layer are frozen). SLidR was used as the control group.}%title
\centering
\scalebox{0.81}
{\setlength{\tabcolsep}{2pt}
\begin{tabular}{c|ccc}% four columns
\toprule
% \begin{tabular}{@{}c@{}}use label\\proportion\end{tabular}
Pre-train Method & Annotation-free & 1\% Fine-tuning 
& \begin{tabular}{@{}c@{}}Pre-training\\Time (h/epoch)\end{tabular}\\
% &\\
\midrule
SLidR & 38.45 & 38.30 & 2.82 \\
TMP(w/ 2D backbone) & 40.12 & 46.80 & 3.44 \\
TMP(w/o 2D backbone) & 40.92 & 46.61 & 2.10 \\
\midrule
\end{tabular}
}
\label{tab:pregeneration}
\end{table}

Finally, we obtain the labels $L_{\mathcal{P}}^{AFI}=\operatorname{argmax}(f_{p_{n}}^{\prime \prime})$ at the final layer of the decoder.

\section{B: Ablation Study}
\label{Appendix_E:ablation_study}

For the ablation study on different setups of AFOV (Tab. \ref{tab:anno-free ablation}), we utilized MaskCLIP \citep{MaskCLIP} as the control group, and FC-CLIP \citep{fcclip}, CAT-Seg \citep{CAT-Seg}, and SAN \citep{san} as the experimental groups. Several sets of experiments were conducted while controlling for irrelevant variables. Meanwhile, for a deeper understanding of the impact of certain ablation targets, we conducted separate experiments employing CAT-Seg as the teacher model, as illustrated in the Tab. \ref{tab:catseg ablation} and Tab. \ref{tab:pregeneration}.

\subsubsection{Performance Gap in Transferring 2D Open-Vocabulary Segmentation Models to 3D Tasks} As shown in Tab. \ref{tab:anno-free ablation}, when conducting ablation experiments with different experimental groups, SAN consistently outperforms FC-CLIP and CAT-Seg under any configuration, achieving a 44.16\% mIoU in the AFOV-baseline, which is 5.16\% mIoU and 5.71\% mIoU higher than the latter two. We record the accuracy (Acc) of different 2D open-vocabulary segmentation models in generating point cloud pseudo-labels in Tab. \ref{tab:ACC}, while also providing their performance on open-vocabulary segmentation in COCO \citep{coco}. We believe that the reasons different teacher models affect student models performance are 1) the ability to comprehend unknown classes, which can affect the accuracy of pseudo-labels, as the number of unknown classes in the class dictionary $\mathcal{C}$ is far greater than that of known classes; Meanwhile, 2) the degree of alignment between image and text features, with SAN exhibiting the largest disparity, making the fitting of the objective loss Eq. \ref{L-I-P},\ref{L-T-P1} more challenging during TMP. It is worth noting that models trained with FC-CLIP and CAT-Seg, even after TMP and AFI, do not outperform the baseline of SAN on annotation-free segmentation. This indicates that \textbf{A Great Teacher Model is All You Need}.

\subsubsection{The Role of Class Dictionary} The class dictionary serves to bridge the datasets' stuff-classes with open vocabularies. Directly inputting semantic classes of the dataset as prompts into 2D open-vocabulary segmentation models is inefficient, as open-vocabulary models are better suited to understand prompts with smaller semantic granularity. As observed in Tab. \ref{tab:catseg ablation}, not using a class dictionary resulted in a -6.14\% mIoU decrease compared to using one.

\subsubsection{Ablation Targets in TMP} In Tab. \ref{tab:catseg ablation}, employing CAT-Seg \citep{CAT-Seg} as the teacher model, we investigate the efficacy of TMP through two ablation experiments: 1) replacing image features with the outputs of self-supervised ResNet50 \citep{resnet} backbone using MoCov2 \citep{mocov2_1,mocov2_2}, and 2) excluding text-point contrastive loss $\mathcal{L}_{T-P}$. Results from fine-tuning with 1\% data and annotation-free point cloud segmentation experiments demonstrate the superiority of TMP over both 1) and 2). Moreover, we affirm the positive impact of TMP on the task of annotation-free point cloud segmentation. As we can see from Tab. \ref{tab:anno-free ablation}, incorporating TMP into baselines of FC-CLIP, CAT-Seg, and SAN improved their respective scores by 2.7\% mIoU, 2.47\% mIoU, and 3.26\% mIoU, compared to pre-training with SLidR \citep{SLidR}.

\subsubsection{The Role of Superpoint-superpixel in TMP} Superpoint and superpixel can avoid self-conflict among point groups (such as $k$-NN groups) with the same semantics. In Tab. \ref{tab:catseg ablation}, $SP$ represents whether superpixel-superpoint is used; otherwise, we use $k$-NN to divide up point cloud. The experimental results show that using superpixel-superpoint leads to performance improvements of +1.58\% mIoU in the annotation-free segmentation and +4.74\% mIoU in the 1\% fine-tuning task.

\subsubsection{Feature Pre-generation} In the experiment of adding a 2D backbone in Tab. \ref{tab:pregeneration}, similar to SLidR \citep{SLidR}, we only activated the output layer of the added 2D backbone (with other layers frozen). The results show that TMP (w/o 2D backbone) has a 25\% and 39\% improvement in runtime over SLidR and TMP (w/ 2D backbone). Moreover, adding a 2D network does not provide performance advantages to TMP. At the same time, the storage burden of pre-generated features is negligible (<7GB). After considering all factors, we retained the step of pre-generating 2D features.

\subsubsection{Effect of AFI} AFI leverages 3D spatial information to enhance the segmentation of unannotated point clouds. Employing AFI on models of FC-CLIP, CAT-Seg, and SAN reached 41.89\%, 41.79\%, and 46.26\% mIoU, a boost of 2.89\%, 3.34\%, and 2.1\% mIoU over their baseline (Tab. \ref{tab:anno-free ablation}). It is noteworthy that the combined usage of TMP and AFI results in a greater improvement compared to using either one alone. However, this improvement is not linearly additive.

\subsubsection{The Role of Pseudo-Label Guided Knowledge Distillation in Fine-tuning Tasks} It is important to note that if we treat AFOV as a pre-training method, it does not utilize the validation set in either stage. The two-stage AFOV outperforms the one-stage AFOV-TMP for downstream tasks on nuScenes, as shown in Tab. \ref{tab:downstream Contrastive}. The superiority arises from the second-stage annotation-free training and the downstream tasks employing the same loss function. Essentially, this is over-fitting on the nuScenes data, which is particularly effective during linear probing. However, annotation-free training weakens the model's generalization ability, resulting in inferior performance when transferred to other datasets, \textit{e.g.}, downstream tasks on SemanticKITTI (pre-trained on nuScenes) compared to using only TMP.

\begin{figure*}[tb]
  \centering
  \includegraphics[width=\linewidth]{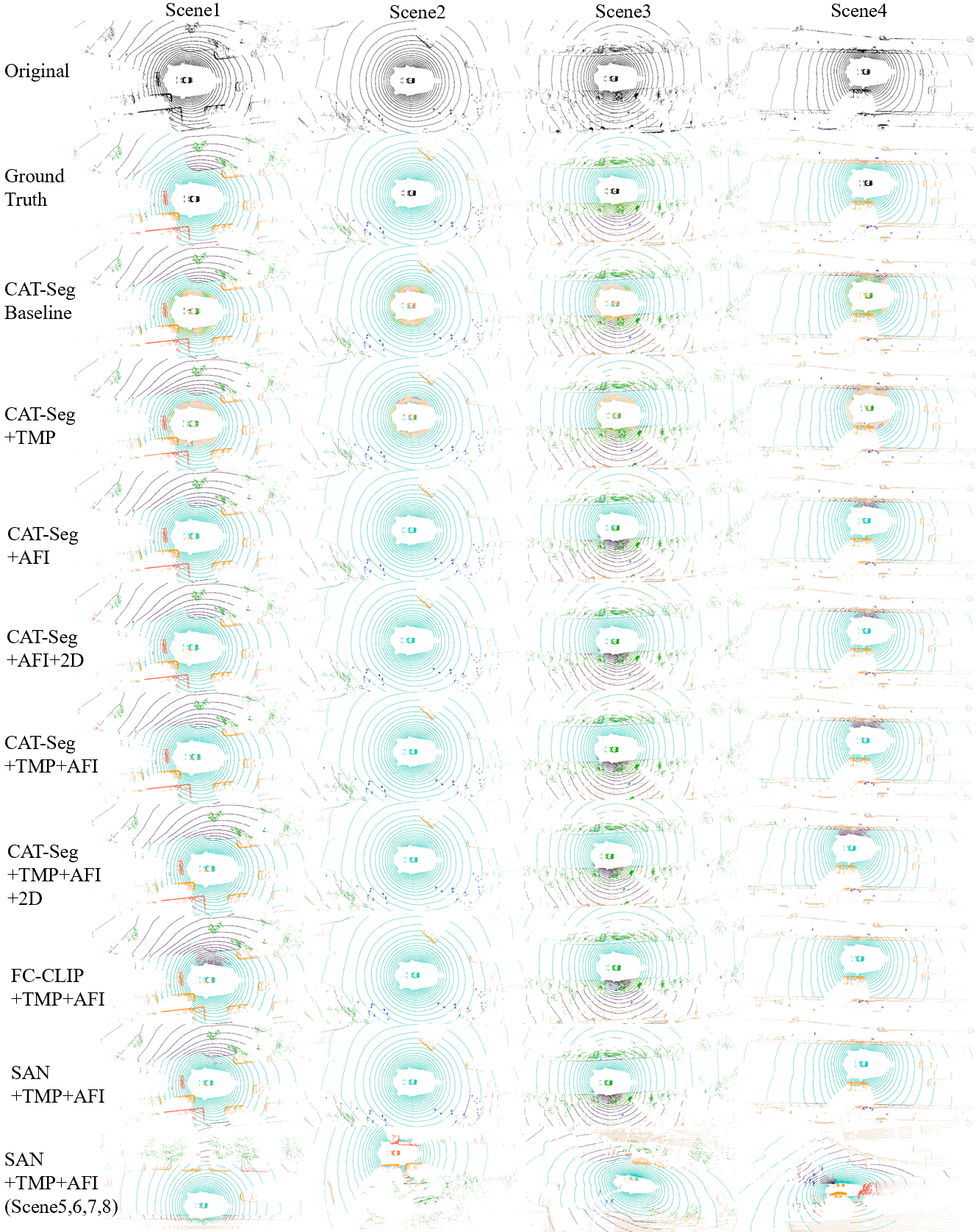}
  \caption{Annotation-free segmentation demonstrations of varied configurations on nuScenes. Here we can intuitively perceive the improvements TMP and AFI bring to the model. Simultaneously, we observe that the optimal outcome (SAN+TMP+AFI) closely approximates the Ground Truth. +2D denotes the utilization of image assistance in AFI.}
  \label{Appendix_B:performance_demo}
\end{figure*}

We present additional annotation-free segmentation results in Fig. \ref{Appendix_B:performance_demo}.

\section{C: Liminations}
\label{Appendix_D:liminations}

Our framework can utilize image assistance during annotation-free segmentation, which is accomplished through post-processing with pseudo labels generated by 2D open-vocabulary models. However, our image assistance offers limited improvement (an average increase of 0.5\% mIoU). Moreover, this work primarily addresses annotation-free learning tasks, wherein pseudo-labels are employed during training, resulting in a relatively weak understanding of point clouds. Besides, due to the specificity of the class dictionary and AFI, it may only be suitable for segmenting outdoor point clouds.

% \section{f}
% \label{Appendix_D:f}
% This task establishes a new baseline for unannotated point cloud segmentation and is hoped to contribute to more in-depth 2D-3D transfer learning in the future. Additionally, since many 2D-3D fusion models are more competitive than single-modal models, exploring methods that complete multimodal perception through feature fusion is worthwhile. We look forward to further development in unannotated perception tasks in the future.

\section{D: Attached Diagrams}

\begin{figure*}[tb]
  \centering
  \includegraphics[width=\linewidth]{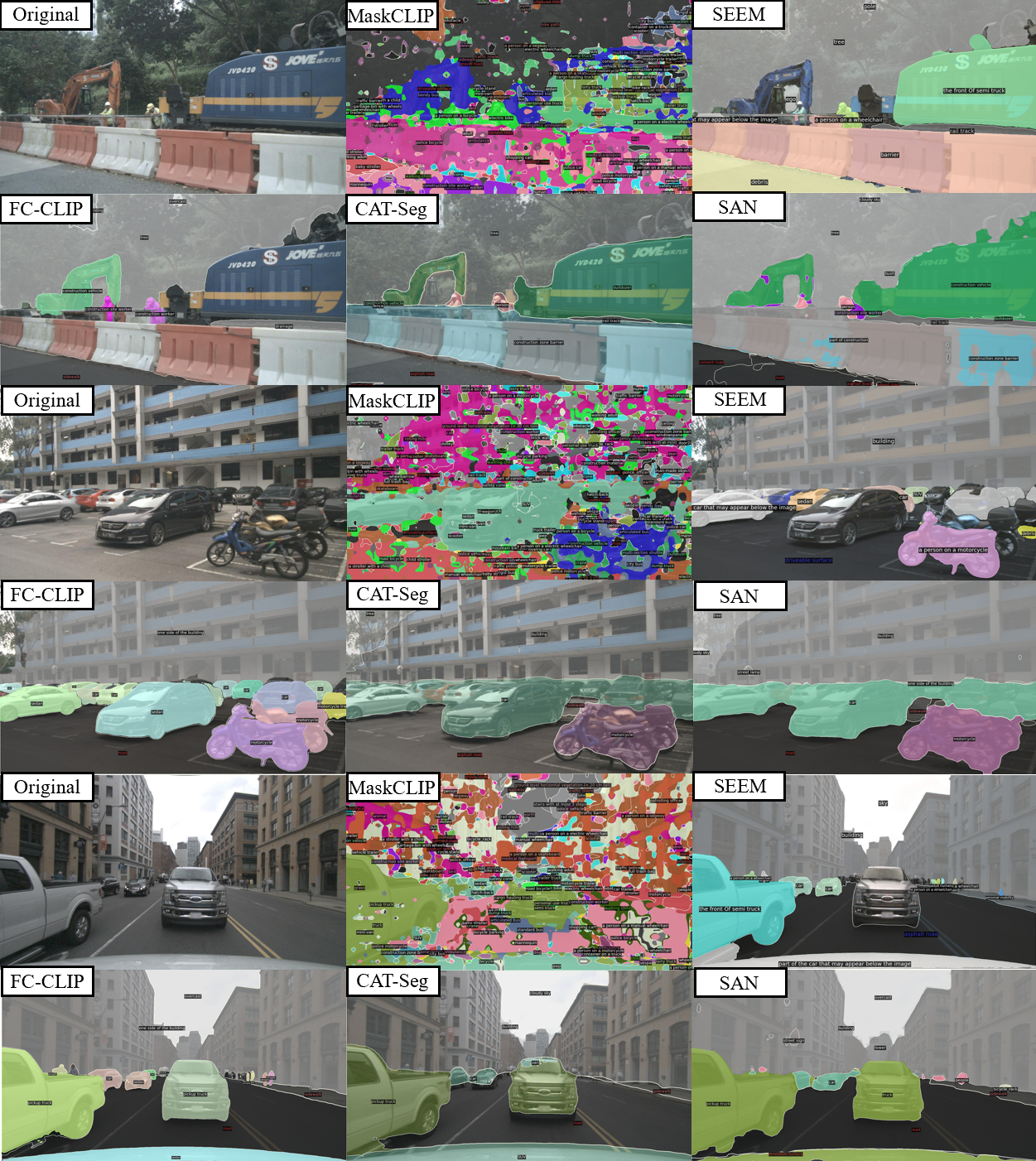}
  \caption{More 2D open-vocabulary segmentation results of different models on nuScenes. }%The open-vocabulary segmentation results of SEEM using the class dictionary $\mathcal{C}$ did not meet our expectations, including annotated errors in categories. Consequently, we abandon using SEEM as the teacher model.}
  \label{Appendix_C:ovseg_result}
\end{figure*}

\clearpage

\end{document}